\documentclass[10pt]{article}
\usepackage[margin=3cm]{geometry}
\usepackage{amssymb}
\usepackage{amsmath}
\usepackage{graphicx}
\usepackage{subfigure}
\usepackage{itmath}
\usepackage{color}
\usepackage{comment}
\usepackage{bussproofs}
\usepackage{times}
\usepackage{makecell}

\newtheorem{definition}{Definition}
\newtheorem{proposition}{Proposition}

\newcommand{\aspicplus}{\textsc{aspic$^\mathsf{+}$}}
\newcommand{\aspicplust}{ASPIC$^\mathsf{+}$}
\newcommand{\aspic}{\textsc{aspic}}
\newcommand{\aspicminus}{\textsc{aspic$_{D}^\mathsf{+}$}}

\newcommand{\args}[1]{\ensuremath{\mc{A}}(#1)}

\newcommand{\mc}[1]{\mathcal{#1}}

\newcommand{\conc}{\ensuremath{\mathtt{Conc}}}
\newcommand{\concs}{\ensuremath{\mathtt{Concs}}}
\newcommand{\sub}{\ensuremath{\mathtt{Sub}}}
\newcommand{\prem}{\ensuremath{\mathtt{Prem}}}
\newcommand{\just}{\ensuremath{\mathtt{Just}}}
\newcommand{\toprule}{\ensuremath{\mathtt{TopRule}}}

\newcommand{\lang}{\ensuremath{\mathcal{L}}}
\newcommand{\rules}{\ensuremath{\mathcal{R}}}

\newcommand{\To}{\Rightarrow}
\newcommand{\vsim}{\mathrel{|\kern-.4em\sim}}
\newcommand{\vsimarg}{\vsim_{a}}
\newcommand{\vsimj}{\vsim_{j}}

\newenvironment{proof}{\vskip 10pt\noindent\textbf{Proof}}{\hfill$\Box$}

\newcommand{\shortcite}[1]{\cite{#1}}

\newcommand\T{\rule{0pt}{2.6ex}}
\newcommand\B{\rule[-1.2ex]{0pt}{0pt}}

\newcommand{\Defeats}{\ensuremath{\mathit{Defeats}}}

\newenvironment{bprooftree}
  {\leavevmode\hbox\bgroup}
  {\DisplayProof\egroup}

\begin{document}
\title{On the links between argumentation-based reasoning and nonmonotonic reasoning}

\author{Zimi Li \\
Department of Computer Science \\
Graduate Center, City University of New York\\ 
E-mail: zli2@gradcenter.cuny.edu
\and Nir Oren \\
Department of Computing Science \\ 
University of Aberdeen\\ 
E-mail: n.oren@abdn.ac.uk
\and Simon Parsons \\
Department of Informatics \\ 
King's College London\\ 
E-mail: simon.parsons@kcl.ac.uk}

\date{7th September 2016}

\maketitle
\begin{abstract}
In this paper we investigate the links between instantiated argumentation systems and the axioms for non-monotonic reasoning described in \cite{kraus1990nonmonotonic} with the aim of characterising the nature of argument based reasoning. In doing so, we consider two possible interpretations of the consequence relation, and describe which axioms are met by \aspicplus\ under each of these interpretations.  We then consider the links between these axioms and the rationality postulates. Our results indicate that argument based reasoning as characterised by \aspicplus is --- according to the axioms of  \cite{kraus1990nonmonotonic} --- non-cumulative and non-monotonic, and therefore weaker than the weakest non-monotonic reasoning systems they considered possible. This weakness underpins \aspicplus's success in modelling other reasoning systems, and we  conclude by considering the relationship between \aspicplus and other weak logical systems.
\end{abstract}

\section{Introduction}
\label{sec:intro}
The rationality postulates proposed by Caminada and Amgoud \shortcite{caminada2007evaluation} have been influential in the development of instantiated argumentation systems. These postulates identify desirable properties for the conclusions drawn from an argument based reasoning process, and focus on the effects of non-defeasible rules within an argumentation system. However, these postulates provide no desiderata with regards to the conclusions drawn from the defeasible rules found within an argumentation system. This latter type of rule is critical to argumentation, and identifying postulates for such rules is therefore important.
At the same time, a large body of work exists which deals with non-monotonic reasoning (NMR). Such NMR systems (exemplified by approaches such as circumscription \cite{mccarthy1980circumscription}, default logic \cite{reiter1980logic} and auto-epistemic logic \cite{moore1985semantical}) introduce various approaches to handling defeasible reasoning, and axioms have been proposed to categorise such systems \cite{kraus1990nonmonotonic}. 

In this paper we seek to combine the rich existing body of work on NMR with structured argumentation systems. We aim to identify what axioms structured argument systems, exemplified by \aspicplus \cite{modgil2012general}  meet\footnote{\aspicplus\ was selected for this study due to its popularity, and its ability to model a variety of other structured systems \cite{DBLP:journals/argcom/ModgilP14}.}. In doing so, we also wish to investigate the links between NMR axioms and the rationality postulates. Doing so may, in the future, allow us to identify additional postulates which apply to the defeasible portion of structured argumentation systems. 

\section{\aspicplust\ Argumentation Framework}
\label{sec:aspicplus}
\aspicplus\ \cite{modgil2012general} is a widely used formalism for structured argumentation, which satisfies the rationality postulates of \cite{caminada2007evaluation}. Arguments within \aspicplus\ are constructed by chaining two types of inference rules, beginning with elements of a knowledge base. The first type of inference rule is referred to as a  \emph{strict} rule, and represents rules whose conclusion can unconditionally be drawn from a set of premises. This is in contrast to \emph{defeasible} inference rules, which allow for a conclusion to be drawn from a set of premises as long as no exceptions or contrary conclusions exist. 
\begin{definition}
An \emph{argumentation system} is a triple $AS = \langle \lang, \rules, n \rangle$ where:
\begin{itemize}
\item $\lang$ is a logical language closed under negation\footnote{This is a small deviation from the standard presentation of \aspicplus, which makes use of a contrariness function which need not be symmetrical.} $\bar{\cdot}$.
\item $\rules = \rules_s \cup \rules_d$ is a set of strict ($\rules_s$) and defeasible ($\rules_d$) inference rules of the form $\phi_1, \ldots, \phi_n \to \phi$ and $\phi_1, \ldots, \phi_n \To \phi$ respectively (where $\phi_i, \phi$ are meta-variables ranging over wff in $\lang$), and $\rules_s \cap \rules_d = \emptyset$.
\item $n : \rules_d \mapsto \lang$ is a naming convention for defeasible rules.
\end{itemize}
\end{definition}
We write $\phi_1, \ldots, \phi_n \leadsto \phi$ if $\rules$ contains a rule of the form $\phi_1, \ldots, \phi_n \to \phi$ or $\phi_1, \ldots, \phi_n \To \phi$.
\begin{definition}
A \emph{knowledge base} in an argumentation system $\langle \lang, \rules, n \rangle$ is a set $\mc{K} \subseteq \lang$ consisting of two disjoint subsets $\mc{K}_n$ (the axioms) and $\mc{K}_p$ (the ordinary premises).
\end{definition}
An argumentation theory consists of an argumentation system and knowledge base.
\begin{definition}
An \emph{argumentation theory} $AT$ is a pair $\langle AS, \mc{K} \rangle$, where $AS$ is an argumentation system $AS$ and  $\mc{K}$ is a knowledge base.
\end{definition}
An argumentation theory is \emph{strict} iff $\mc{R}_d = \emptyset$ and $\mc{K}_p=\emptyset$, and is \emph{defeasible} iff it is not strict.

To ensure that reasoning meets norms for rational reasoning according to the \emph{rationality postulates} of \cite{caminada2007evaluation}, an \aspicplus\ argumentation system must be such that its strict rules are closed under transposition. That is, given a strict rule with premises $\varphi=\{\phi_1, \ldots,\phi_n\}$ and conclusion $\phi$ (written $\varphi \to \phi$), a set of $n$ additional rules of the following form must be present in the system: $\{ \overline{\phi} \} \cup \varphi \backslash \{\phi_i\} \to \overline{\phi_i}$ for all $1 \leq i \leq n$. 

Arguments are defined recursively in terms of sub-arguments and through the use of several functions: $\prem(A)$ returns all the premises of argument $A$; $\conc(A)$ returns $A$'s conclusion, and $\toprule(A)$ returns the last rule used within the argument. $\sub(A)$ returns all of $A$'s sub-arguments. Given this, arguments are defined as follows.
\begin{definition}
\label{def:argument}
An \emph{argument} $A$ on the basis of an argumentation theory $AT = \langle \langle \lang, \rules, n\rangle, \mc{K}\rangle$ is:
\begin{enumerate}
\item $\phi$ if $\phi \in \mc{K}$ with: \prem($A$) = $\{\phi\}$; \conc($A$) = $\{\phi\}$; \sub($A$) = $\{A\}$; \toprule($A$) = undefined.

\item $A_1, \ldots, A_n \to/\To \phi$ if $A_i$ are arguments such that there respectively exists a strict/defeasible rule $\conc(A_1), \ldots, \conc(A_n) \to/\To \phi$ in $\rules_s$ / $\rules_d$. $\prem(A) = \prem(A_1) \cup \ldots \cup \prem(A_n); \conc(A) = \phi; \sub(A) = \sub(A_1) \cup \ldots \cup \sub(A_n) \cup \{A\}; \toprule(A) = \conc(A_1), \ldots, \conc(A_n) \to/\To \phi$.
\end{enumerate}
\end{definition}
We write $\args{AT}$ to denote the set of arguments on the basis of the theory $AT$, and given a set of arguments $\mathbf{A}$, we write $\concs(\mathbf{A})$ to denote the conclusions of those arguments, that is:
\[
\concs(\textbf{A}) = \{ \conc(A) | A\in\textbf{A}\}
\]
Like other argumentation systems, \aspicplus\ utilises conflict between arguments  --- represented through attacks --- to determine what conclusions are justified.

An argument can be attacked in three ways: on its ordinary premises, on its conclusion, or on its inference rules. These three kinds of attack are called undermining, rebutting and undercutting attacks, respectively.
\begin{definition}
\label{def:attack}
An argument $A$ \emph{attacks} an argument $B$ iff $A$ undermines, rebuts or undercuts $B$, where:
\begin{itemize}
\item $A$ undermines $B$ (on $B'$) iff \conc($A$) $= \overline{\phi}$ for some $B'=\phi \in \prem(B)$ and $\phi \in \mc{K}_p$.
\item $A$ rebuts $B$ (on $B'$) iff \conc($A$) $= \overline{\phi}$ for some $B' \in \sub(B)$ of the form $B''_1, \ldots , B''_2 \To \phi$.
\item $A$ undercuts $B$ (on $B'$) iff \conc($A$) $= \overline{n(r)}$ for some $B' \in \sub(B)$ such that $\toprule(B)$ is a defeasible rule $r$ of the form $\phi_1, \ldots, \phi_n \To \phi$.
\end{itemize}
\end{definition}
Note that, in \aspicplus\ rebutting is \emph{restricted}. That is an argument with a strict \toprule\ can rebut an argument with a defeasible \toprule, but not vice versa. (\cite{caminada2014preferences} and \cite{li2015argumentation} introduce the \aspic- and \aspicminus\ systems which use unrestricted rebut). Finally,  A set of arguments is said to be \emph{consistent} iff there is no attack between any arguments in the set.

Attacks can be distinguished by whether they are preference-dependent (rebutting and undermining) or preference-independent (undercutting). The former succeed only when the attacker is preferred. The latter succeed whether or not the attacker is preferred. Within \aspicplus\ preferences over defeasible rules and ordinary premises are combined or lifted to obtain a preference ordering over arguments \cite{modgil2012general}. Here, we are not concerned about the means of combination, but, following \cite{modgil2012general}, we only consider \emph{reasonable} orderings. For our purposes, a reasonable ordering is one such that adding a strict rule or axiom to an argument will neither increase nor decrease its preference level.
\begin{definition}
\label{def:ordering}
 A preference ordering $\preceq$ is a binary relation over arguments, i.e., $\preceq\ \subseteq \mc{A} \times \mc{A}$, where $\mc{A}$ is the set of all arguments constructed from the knowledge base in an argumentation system.
\end{definition}
 Combining these elements results in the following:
\begin{definition}
A \emph{structured argumentation framework} is a triple $\langle \mc{A}, att, \preceq \rangle$, where $\mc{A}$ is the set of all arguments constructed from the argumentation system, $att$ is the attack relation, and $\preceq$ is a preference ordering on $\mc{A}$.
\end{definition}
Preferences over arguments interact with attacks such that \emph{preference-dependent} attacks succeed when the attacking argument is preferred. In contrast \emph{preference-independent} attacks always succeed. Attacks that succeed are called \emph{defeats}.
Using Definition~\ref{def:argument} and the notion of defeat, we can instantiate an abstract argumentation framework from a structured argumentation framework. 
\begin{definition}
An \emph{(abstract) argumentation framework} $AF$ corresponding to a structured argumentation framework $SAF = \langle \mc{A}, att, \preceq \rangle$ is a pair $\langle \mc{A}, \Defeats \rangle$ such that $\Defeats$ is the defeat relation on $\mc{A}$ determined by $SAF$.
\end{definition}
This abstract argumentation framework can be evaluated using standard argumentation semantics \cite{phan1995acceptability}, defining the notion of an extension:
\begin{definition}
Let $AF=\langle \mc{A}, \Defeats \rangle$ be an abstract argumentation framework, Let $A \in \mc{A}$ and $E \subseteq \mc{A}$. $E$ is said to be  \emph{conflict-free} iff there do not exist a $B, C \in E$ such that $B$ defeats $C$. $E$ is said to defend $A$ iff for every $B \in \mc{A}$ such that $B$ defeats $A$, there exists a $C \in E$ such that $C$ defeats $B$. The characteristic function $F : 2^\mc{A} \to 2^\mc{A}$ is defined as $F(E)=\{ A \in \mc{A} | E \mathit{\ defends\ } A\}$. $E$ is called (1) an \emph{admissible} set iff $E$ is conflict-free and $E \subseteq F(E)$; (2) a \emph{complete} extension iff $E$ is conflict-free and $E = F(E)$; (3) a \emph{grounded} extension iff $E$ is the minimal complete extension; and (4) a \emph{preferred} extension iff $E$ is a maximal complete extension, where minimality and maximality are w.r.t. set inclusion.
\end{definition}
We note in passing that other extensions have been defined and refer the reader to  \cite{baroni2011introduction} for further details.
\begin{definition}
If $AF=\langle \mc{A}, Defeats \rangle$ is an abstract argumentation framework, and $E$ is the extension under one of Dung's semantics. We write $\just(\mc{A})$ to denote the set of justified conclusions of the set of arguments $\mc{A}$, i.e., $
\just(\mc{A})=\{ \concs(E) \}
$.
\end{definition}

\section{Axiomatic Reasoning and \aspicplust}
\label{sec:cumulative:axioms}
Kraus \emph{et al.} \cite{kraus1990nonmonotonic}, building on earlier work by Gabbay \shortcite{gabbay:nato}, identified a set of axioms which characterise non-monotonic inference in logical systems, and studied the relationships between sets of these axioms. Their goal was to characterise different kinds of reasoning; to pin down what it means for a logical system to be monotonic or non-monotonic; and --- in particular --- to be able distinguish between the two. Table \ref{table:axioms}  presents the axioms of \cite{kraus1990nonmonotonic}, which we will use to characterise reasoning in \aspicplus. The symbol $\vsim$ encodes a consequence relation, while $\models$ identifies the statements obtainable from the underlying theory. Note that we have altered some of the symbols used in \cite{kraus1990nonmonotonic} to avoid confusion with the notation of \aspicplus. Equivalence is denoted  $\equiv$ (rather than $\leftrightarrow$), and $\hookrightarrow$ (rather than $\rightarrow$) denotes the existence of a rule, acknowledging the fact that a rule may be strict or defeasible.
\begin{table}
\begin{tabular}{|l|c|l|}
\hline
\textbf{Abbr.} & \textbf{Axiom} & \textbf{Name} \\
\hline
Ref &
$
\alpha \vsim \alpha
$ & Reflexivity \\
\hline

LLE &
\begin{bprooftree}
\AxiomC{$\T\models \alpha \equiv \beta$}
\AxiomC{$\alpha \vsim \gamma$}
\BinaryInfC{$\beta \vsim \gamma$\B}
\end{bprooftree}
& \makecell[l]{Left Logical\\Equivalence} \\
\hline

RW & \T
\begin{bprooftree}
\AxiomC{\T$\models \alpha \hookrightarrow \beta$}
\AxiomC{$\gamma \vsim \alpha$}
\BinaryInfC{$\gamma \vsim \beta$\B}
\end{bprooftree}
& \makecell[l]{Right\\Weakening} \\
\hline

Cut & 
\begin{bprooftree}
\AxiomC{$\T\alpha \wedge \beta \vsim \gamma$}
\AxiomC{$\alpha \vsim \beta$}
\BinaryInfC{$\alpha \vsim \gamma$\B}
\end{bprooftree}
& \\
\hline

CM & 
\begin{bprooftree}
\AxiomC{\T$\alpha \vsim \beta$}
\AxiomC{$\alpha \vsim \gamma$}
\BinaryInfC{$\alpha \wedge \beta \vsim \gamma$\B}
\end{bprooftree}
& \makecell[l]{Cautious\\Monotonicity} \\
\hline

M & \begin{bprooftree}
\AxiomC{\T$\models \alpha \hookrightarrow \beta$}
\AxiomC{$\beta \vsim \gamma$}
\BinaryInfC{$\alpha \vsim \gamma$\B}
\end{bprooftree} 
& Monotonicity \\
\hline

T & \begin{bprooftree}
\AxiomC{\T$\alpha \vsim \beta$}
\AxiomC{$\beta \vsim \gamma$}
\BinaryInfC{$\alpha \vsim \gamma$\B}
\end{bprooftree}
& Transitivity \\
\hline

CP & \begin{bprooftree}
\AxiomC{\T$\alpha \vsim \beta$}
\UnaryInfC{$\overline{\beta} \vsim \overline{\alpha}$\B}
\end{bprooftree}
& Contraposition \\
\hline
\end{tabular}
\caption{The axioms from \cite{kraus1990nonmonotonic} that we will consider.}
\label{table:axioms}
\end{table}

Consequence relations that satisfy Ref, LLE, RW, Cut and CM are said to be \emph{cumulative}, and \cite{kraus1990nonmonotonic} describes them as being the weakest interesting logical system. Cumulative consequence relations which also satisfy CP are \emph{monotonic}, while consequence relations that are cumulative and satisfy M are called \emph{cumulative monotonic}. Such relations are stronger than cumulative but not monotonic in the usual sense.

To determine which axioms \aspicplus does or does not comply with, we must decide how different aspects of the axioms should be interpreted. 
We interpret the consequence relation $\vsim$ in a couple of ways that are natural in the context of \aspicplus --- describing these in detail later --- and which fit with the high level meaning of ``if $\alpha$ is in the knowledge base, then $\beta$ is a conclusion'', or ``$\beta$ is a consequence of $\alpha$''.

Assuming such an interpretation of $\alpha\vsim\beta$ we can consider the meaning of the axioms.
Some axioms are clear.
For example, axiom T says that if $\beta$ is a consequence of $\alpha$, and $\gamma$ is a consequence of $\beta$, then $\gamma$ is a consequence of $\alpha$.
Other axioms are more ambiguous.
Does $\alpha\wedge\beta\vsim\gamma$ in Cut mean that $\gamma$ is a consequence of the conjunction $\alpha\wedge\beta$ or is it a consequence of $\alpha$ and $\beta$ together?
In other words is $\wedge$ a feature of the language underlying the reasoning system, or a feature of the meta-language in which the properties are written?
Similarly, given the distinction between strict and defeasible rules, is $\alpha \hookrightarrow \beta$ a strict rule in \aspicplus, a defeasible rule, or some statement in the property meta-language?

The symbols found in the axioms are interpreted as follows.
\begin{itemize}

 \item $\models\alpha$ means that $\alpha$ is an element of the
   relevant argumentation theory $AT$. 
   
 \item $\alpha\wedge\beta$ means both $\alpha$ and $\beta$, in
   particular in Cut and CM, it means that both $\alpha$ and $\beta$ are in the knowledge base.

 \item $\alpha\equiv\beta$ is taken --- as usual --- to abbreviate the formula $(\alpha\hookrightarrow\beta)\wedge(\beta\hookrightarrow\alpha)$. We assume $\alpha\hookrightarrow\beta$ and $\beta\hookrightarrow\alpha$ have the same interpretation, i.e., both or neither are strict.

 \item $\alpha\hookrightarrow\beta$ has two interpretations. We have the \emph{strict} interpretation in which $\alpha\hookrightarrow\beta$ denotes a strict rule $\alpha\to\beta$ in \aspicplus, and the \emph{defeasible} interpretation in which $\alpha\hookrightarrow\beta$ denotes either a strict or defeasible rule. 
 We denote the latter interpretation by writing $\alpha\leadsto\beta$.   
\end{itemize}

\section{Axioms and Consequences in \aspicplust}
\label{sec:cumulative:aspicplus}

We now consider the two different interpretations of the non-monotonic consequence relation $\vsim$ described above, identifying which axioms each interpretation satisfies. In the following, we assume an arbitrary \aspicplus\ argumentation theory $AT = \langle \langle \lang, \rules, n\rangle, \mc{K}\rangle$, and write $AT_x$ for an extension of this augmentation theory also containing proposition $x$: $AT_x= \langle \langle \lang, \rules, n\rangle, \mc{K}\cup \{x\}\rangle$. An argument present in the latter, but not former, theory is denoted $A^x$.

\subsection{Argument Construction}
\label{sec:argument}

We begin by considering the consequence relation as representing argument construction. I.e., we interpret $\alpha \vsim \beta$  as meaning that if $\alpha$ is in the 
axioms or ordinary premises of a theory, we can construct an argument for $\beta$.
More precisely:
\begin{definition}
We write $\alpha\vsimarg\beta$, if for every \aspicplus\ argumentation theory $AT = \langle \langle \lang, \rules, n\rangle, \mc{K}\rangle$ such that $\beta\not\in \concs(\args{AT})$, it is the case that $\beta\in \concs(\args{AT_\alpha})$. 
\end{definition}
\begin{proposition}
The consequence relation $\vsimarg$ is cumulative for both strict and defeasible theories.
\begin{proof}
Consider an arbitrary theory $AT= \langle \langle \lang, \rules, n\rangle, \mc{K}\rangle$. 
\textbf{[Ref]} Given a theory  $AT_\alpha$, we have an argument $A^\alpha = [\alpha]$, so Ref holds for $\vsimarg$.
\textbf{[LLE]} Since $\alpha\vsimarg\gamma$, $AT_\alpha$ contains a chain of arguments $A^\alpha_1, A^\alpha_2, \ldots, A^\alpha_n$ with $A^\alpha_1=[\alpha]$ and $\conc(A^\alpha_n)=\gamma$. Given $\models\alpha\equiv\beta$, we have that both $\alpha\leadsto\beta$ and $\beta\leadsto\alpha$ are in the theory $AT$, so in the theory $AT_\beta$. In the theory $AT_\beta$, we obtain a chain of arguments $B^\beta_0=[\beta], B^\beta_1=[B^\beta_0 \leadsto\alpha], A^\beta_2, \ldots, A^\beta_n$. That is $\beta\vsimarg\gamma$. Therefore, both strict and defeasible versions of LLE hold for $\vsimarg$.
\textbf{[RW]} Since $\gamma \vsimarg \alpha$  in theory $AT_\gamma$, there is a chain of arguments $A^\gamma_1, A^\gamma_2, \ldots, A^\gamma_n$ with $A^\gamma_1=[\gamma]$ and $\conc(A^\gamma_n)=\alpha$. Given $\models \alpha\hookrightarrow\beta$, $\alpha\leadsto\beta$ is in the theory $AT$, it is also in  $AT_\gamma$. In $AT_\gamma$, we have a chain of arguments $A^\gamma_1, \ldots, A^\gamma_n, A^\gamma_{n+1}=[A^\gamma_n\To\beta]$. Thus, $\gamma\vsimarg\beta$, and both strict and defeasible versions of RW hold for $\vsimarg$.
\textbf{[Cut]} Since $\alpha \wedge \beta \vsimarg \gamma$, in theory $AT_{\alpha,\beta}$, there is a chain of arguments $A^{\alpha,\beta}_1, A^{\alpha,\beta}_2, \ldots, A^{\alpha,\beta}_n$ with $A^{\alpha,\beta}_1=[\alpha]$, $A^{\alpha,\beta}_2=[\beta]$ and $\conc(A^{\alpha,\beta}_n)= \gamma$. In the theory $AT_\alpha$, since $\alpha \vsimarg \beta$, there is a chain of arguments $B^\alpha_1, B^\alpha_2, \ldots, B^\alpha_m$ with $B^\alpha_1=[\alpha]$ and $\conc(B^\alpha_m)=\beta$. There is also a chain of arguments $B^\alpha_1, B^\alpha_2, \ldots, B^\alpha_m, A^\alpha_3, \ldots, A^\alpha_n$. That is $\alpha\vsimarg\gamma$. Therefore, cut holds for $\vsimarg$.
\textbf{[CM]} Since $\alpha \vsimarg \gamma$  $AT_\alpha$ has a chain of arguments $A^\alpha_1, \ldots, A^\alpha_n$ with $A^\alpha_1=[\alpha]$ and $\conc(A^\alpha_n)=\gamma$. $AT_{\alpha,\beta}$ has a similar chain of arguments $A^{\alpha,\beta}_1, \ldots, A^{\alpha,\beta}_n$, so $\alpha\wedge\beta \vsimarg \gamma$. CM thus holds for $\vsimarg$.

Since all of the above axioms hold, $\vsimarg$ is cumulative for both strict and defeasible theories.
\end{proof}
\end{proposition}
\begin{proposition}
The consequence relation $\vsimarg$ satisfies M and T for both strict and defeasible theories.
\begin{proof}
Consider an arbitrary theory $AT= \langle \langle \lang, \rules, n\rangle, \mc{K}\rangle$. 
\textbf{[M]} Since $\beta\vsimarg\gamma$, in the theory $AT_\beta$, there is a chain of arguments $A^\beta_1, A^\beta_2, \ldots, A^\beta_n$ with $A^\beta_1=[\beta]$ and $\conc(A^\beta_n)=\gamma$. Given $\models \alpha\hookrightarrow\beta$, we have $\alpha\leadsto\beta$ in the theory $AT$, and also in the theory $AT_\alpha$. In the latter, there is a chain of arguments $B^\alpha_0=[\alpha], B^\alpha_1=[B^\alpha_0 \leadsto \beta], A^\alpha_2, \ldots, A^\alpha_n$. That is $\alpha\vsimarg\gamma$. Therefore, both strict and defeasible versions of M hold for $\vsimarg$.
\textbf{[T]} Since $\beta\vsimarg\gamma$, in  $AT_\beta$, there is a chain of arguments $B^\beta_1, B^\beta_2, \ldots, B^\beta_m$ with $B^\beta_1=[\beta]$ and $\conc(B^\beta_m)=\gamma$. Similarly, since $\alpha\vsimarg\beta$, in  $AT_\alpha$, there is a chain of arguments $A^\alpha_1, A^\alpha_2, \ldots, A^\alpha_n$ with $A^\alpha_1=[\alpha]$ and $\conc(A^\alpha_n)=\beta$. Combined with $B^\alpha_1, B^\alpha_2, \ldots, B^\alpha_m$, we obtain a chain of arguments $A^\alpha_1, A^\alpha_2, \ldots, A^\alpha_n, B^\alpha_2, \ldots, B^\alpha_m$. That is $\alpha\vsimarg\gamma$. Therefore, T holds for $\vsimarg$.
\end{proof}
\end{proposition}
Thus $\vsimarg$ is cumulative monotonic for all theories. It is not, however, monotonic.
\begin{proposition}
\label{prop:atol}
$\vsimarg$ does not satisfy axiom CP.
\begin{proof} Consider this counter-example. $\mc{K}= \{c\}$, $\mc{R}_s = \{\alpha, c \to d; \alpha, \overline{d} \to \overline{c}; c, \overline{d} \to \overline{\alpha}; \alpha \to e; \overline{e} \to \overline{\alpha};  d,e \to \beta; d,\overline{\beta} \to \overline{e}; \overline{\beta}, e \to \overline{d}\}$
We have $\alpha \vsimarg \beta$ but not $\overline{\beta} \vsimarg \overline{\alpha}$. Therefore, CP does not hold for $\vsimarg$.
\end{proof}
\end{proposition}

\subsection{Justified Conclusions}
\label{sec:justified}
Next we consider $\alpha\vsim\beta$ as meaning that if $\alpha$ is in a theory, we can construct an argument for $\beta$ such that $\beta$ is in the set of justified conclusions (regardless of preferences). 
In the following, we will consider an arbitrary extension containing the justified conclusions, these proofs are therefore applicable to any extension based semantics.

\begin{definition}
We write $\alpha\vsimj\beta$, if for every \aspicplus\ argumentation theory $AT = \langle \langle \lang, \rules, n\rangle, \mc{K}\rangle$ such that $\beta\not\in \just(\args{AT})$, it is the case that $\beta\in \just(\args{AT_\alpha})$ 
\end{definition}
It is worth noting the following result.
\begin{proposition}
  If $\alpha\vsimj\beta$ then $\alpha\vsimarg\beta$.
  \begin{proof}
    Follows immediately from the definitions --- for $\beta$ to be a justified
    conclusion, there must first be an argument with $\beta$ as a
    conclusion.
  \end{proof}
  \label{prop:atoj}
\end{proposition}
Since there are, in general, less justified conclusions of a theory than there are arguments, $\vsimj$ is a more restrictive notion of consequence than $\vsimarg$.
It is therefore no surprise to find that fewer of the axioms from \cite{kraus1990nonmonotonic} hold. We have the following.
\begin{proposition}
The consequence relation $\vsimj$ does \emph{not} satisfy reflexivity, or the defeasible versions of LLE and RW.
\begin{proof} 
\textbf{[Ref]} Counterexample: consider an \aspicplus\ theory that contains: $\mc{K}_n=\{\overline{a}\}$ and $\mc{R}=\emptyset$. Here, we have an argument $A=[\overline{a}]$. If  $a$ is in the knowledge base $\mc{K}_p$, we have  another argument $B=[a]$. However, $B$ is defeated by $A$, but not vice versa. So $B$ is not in any extension, and Ref does not hold for $\vsimj$.
\textbf{[LLE (defeasible version)]}
Counter-example: consider an \aspicplus\ theory that contains $\mc{K}_n=\{c\}$ and $\mc{R}=\{a \To b; b \To a; a \To r; c \to \overline{n_1}\}$ where $n(b \To a) = n_1$. Here, $a \vsimj r$, but, $b \not\vsimj r$. Therefore, the defeasible version of LLE does not hold for $\vsimj$.
\textbf{[RW (defeasible version)]} Consider any \aspicplus\ theory that contains $\overline{\beta}$ in its axioms. For such a theory, $\beta$ will not appear in any justified conclusions. Therefore, the defeasible version of RW does not hold for $\vsimj$.
\end{proof}
\end{proposition}
\begin{proposition}
 $\vsimj$ satisfies the strict version of LLE and RW, Cut and CM for strict or defeasible theories.
\begin{proof}
Consider an arbitrary theory $AT= \langle \langle \lang, \rules, n\rangle, \mc{K}\rangle$.
\textbf{[LLE (strict version)]]}
Since  $\models \alpha\equiv\beta$, under the strict interpretation, the rule $\beta\to\alpha$ is in $AT$, $AT_\alpha$ and $AT_\beta$. Since $\alpha \vsimj \gamma$, we know that there is an extension $E_\alpha$ containing $A^\alpha_1, A^\alpha_2, \ldots, A^\alpha_n$ with $A^\alpha_1=[\alpha]$ and $\conc(A^\alpha_n)=\gamma$. Furthermore, there is no attack\footnote{Note that since $\vsimj$ considers any preference ordering, attacks and defeats here are equivalent.}  between $A_i$ ($i=1\ldots n$) and $B^\alpha$, where $B^\alpha$ is an argument in $E_\alpha$. In addition, there is no argument with conclusion $\overline{\beta}$ in $E_\alpha$ since $A^\alpha_1$ is in $E_\alpha$ and there is a strict rule $\alpha\to\beta$\footnote{Due to closure under strict rules.}. 
Now consider theory $AT_\beta$, which has a chain of arguments $A^\beta_0=[\beta], A^\beta_1=[A^\beta_0\to\alpha], A^\beta_2, \ldots, A^\beta_n$, where $\conc(A^\beta_n)=\gamma$. There is an extension $E_\beta=\{A^\beta_0, \ldots, A^\beta_n\} \cup (E_\alpha-\{A^\alpha_1, \ldots, A^\alpha_n\})$ in $AT_\beta$ under the same semantic. Therefore  strict  LLE holds for $\vsimj$.
\textbf{[RW (strict version)]}
Consider the extension $E_\gamma$ in $AT_\gamma$ containing an argument $A^\gamma$ with $\conc(A^\gamma)=\alpha$. Since $\models \alpha \leadsto \beta$, under the strict interpretation, we know that $\alpha\to\beta$ is in $AT_\gamma$. Because of closure under strict rules, $B^\gamma=[A^\gamma\to \beta]$ is in $E_\gamma$. Therefore the strict version of RW holds for $\vsimj$.
\textbf{[Cut]}
Since $\alpha\wedge\beta\vsimj\gamma$, we know that there is an extension $E_{\alpha,\beta}$ of $AT_{\alpha,\beta}$ containing $B^{\alpha,\beta}, A^{\alpha,\beta}_1, \ldots, A^{\alpha,\beta}_n$ with $A^{\alpha,\beta}_1=[\alpha]$, $B^{\alpha,\beta}=[\beta]$ and $\conc(A^{\alpha,\beta}_n)= \gamma$. Now consider the theory $AT_\alpha$. Since $\alpha\vsimj\beta$, there is an extension $E_\alpha$ containing $B^\alpha_1, B^\alpha_2 \ldots, B^\alpha_m$ with $B^\alpha_1=[\alpha]$ and $\conc(B^\alpha_m)=\beta$. The set $E_\alpha \cup (E_{\alpha,\beta}-\{B^{\alpha,\beta}\})$ is an extension in $AT_\alpha$ (c.f., LLE above). Therefore cut holds for $\vsimj$.
\textbf{[CM]}
Since $\alpha\vsimj\beta$, $AT_\alpha$ and $AT_{\alpha,\beta}$ contain similar arguments. Since $\alpha\vsimj\gamma$, there is an extension $E_\alpha$ in $AT_\alpha$ containing $A^\alpha_1, \ldots, A^\alpha_n$ with $A^\alpha_1=[\alpha]$ and $\conc(A^\alpha_n)=\gamma$.  $E_\alpha$ is also an extension in $AT_{\alpha,\beta}$, since $AT_\alpha$ and $AT_{\alpha,\beta}$ contain similar arguments. Therefore CM holds for $\vsimj$.
\label{prop:justified}
\end{proof}
\end{proposition}
\begin{proposition}
The consequence relation $\vsimj$ does \emph{not} satisfy M, T or CP for defeasible theories.
\begin{proof}
We will give counter-examples.
\textbf{[M]} Consider the \aspicplus\ theory that includes
 $\mc{K}_n=\{\overline{a}\}$ and $\mc{R}=\{a \to b; \overline{b} \to \overline{a}; b \To \gamma\}$. Thus,  $b\vsimj\gamma$, however, $a\not\vsimj\gamma$. 
 Therefore, M does not hold for $\vsimj$.
\textbf{[T]} Consider the \aspicplus\ theory which includes $\mc{K}=\emptyset$ and $\mc{R}=\{a \To b; b \To c; c \To r; a \to \overline{c}\}$. Thus, $a\vsimj b$ and $b\vsimj r$, but $a \not\vsimj r$. 
Therefore, T does not hold for $\vsimj$.
\textbf{[CP]} Since contraposition does not hold for $\vsimarg$, by Proposition~\ref{prop:atol} it cannot hold for $\vsimj$.
\end{proof} 
\end{proposition}
If we consider only strict theories, the following holds.
\begin{proposition}
The consequence relation $\vsimj$ satisfies Ref, M and T for strict theories.
\begin{proof}
If the theory is strict, for any argumentation theory, all conclusions are justified. Therefore, for any strict theory, if $\alpha \vsimarg \beta$, then $\alpha \vsimj \beta$.
We know that $\vsimarg$ holds for Ref, M and T, therefore, $\vsimj$ holds for Ref, M and T in a strict theory.
\end{proof}
\label{prop:vsimj:ref}
\end{proposition}
Thus $\vsimj$ is cumulative monotonic for strict theories.


\subsection{Discussion}

\begin{table}[t]
\caption{
Summary of axioms satisfied by the argumentation-based consequence relations. Y indicates that the axiom holds for strict and defeasible theories without constraint; (Y) that the axiom holds for strict theories only; [Y] that the strict version of the axiom holds in a theory of strict and defeasible rules (i.e., the rule mentioned in the axiom has to be strict, but arguments may contain defeasible rules); and N  that the axiom does not hold for any theory. The left-hand set of axioms are those required for cumulativity.
}
\begin{tabular}{lccccc|ccc}
           & Ref & LLE & RW  & Cut & CM &  M   & T & CP  \\
$\vsimarg$ & Y   & Y   & Y   & Y   & Y  &  Y   & Y & N \\
$\vsimj$   & (Y) & [Y] & [Y] & Y   & Y  & (Y) & (Y) & N \\
\end{tabular}
\label{tab:summary}
\end{table}

The results from the previous sections are summarized in Table~\ref{tab:summary}.
What light does the above shine on \aspicplus\ and argumentation-based reasoning in general?
Considering $\vsimarg$, it is no surprise that the relation is cumulative monotonic and satisfies the axiom M which captures a form of monotonicity.
It is clear from the detail of \aspicplus, and indeed any argumentation system, that the number of arguments grows over time.
However, the fact that $\vsimarg$ is not monotonic in the same strict sense as classical logic, and so is strictly weaker, as a result of not satisfying CP, is a bit more interesting.
This is, of course, because arguments are not subject to the law of the excluded middle --- it is perfectly possible for there to be arguments for $\alpha$ and $\overline{\alpha}$ from the same theory.

Turning to $\vsimj$, this is perhaps a more reasonable notion of consequence for \aspicplus\ than $\vsimarg$. If $\alpha\vsimj\beta$, then there is an argument for $\beta$ which holds despite any attacks (in the scenario we have considered, where all attacks may be defeats for some preference ordering --- and therefore succeed --- there can still be attacks on the argument for $\beta$, but the attacking arguments must themselves be defeated).
This is quite a restrictive notion of consequence in a representation that allows for conflicting information, and as Table~\ref{tab:summary} makes clear, $\vsimj$ is a relatively weak notion of consequence. It obeys less of the axioms and thus sanctions less conclusions than the non-monotonic logics analysed in \cite{kraus1990nonmonotonic}, for example.
For defeasible theories $\vsimj$ is not cumulative, and only satisfies LLE and RW if the rules applied in those axioms are strict.
As we pointed out above, at the time that \cite{kraus1990nonmonotonic} was published, cumulativity was considered the minimum requirement of a useful logic\footnote{This position was doubtless a side-effect of the fact that at that time there were no logics that did not obey cumulativity. Subsequent discovery of logics of causality that are not cumulative suggests that this view should  be revised.}. Whether or not one accepts this, it is clear that \aspicplus\ is weak. 
But is it too weak?
To answer this, we should consider the cause of the weakness, which as Table~\ref{tab:summary} shows is due to LLE, RW and Ref.

LLE and RW only hold if the axioms are only applied to strict rules. 
In both cases, the effect of the axiom is to extend an existing argument, either switching one premise for another (LLE), or adding a rule to the conclusion of an argument (RW).
While having these axioms hold for defeasible rules would allow $\vsimj$ to be cumulative for defeasible theories, it is not reasonable for justified conclusions to be drawn under these circumstances --- using LLE or RW to extend arguments with defeasible rules by definition means that the new arguments can be defeated.
Thus their conclusions may not be justified.

A similar argument applies to Ref.
Proposition~\ref{prop:vsimj:ref} tells us that Ref holds for $\vsimj$ for strict theories. 
In effect that means that $\alpha$ has to be an axiom.
If Ref were to hold for defeasible theories, $\alpha$ could be a premise. 
But premises can be defeated, again by definition, so it is not appropriate to directly conclude that any premise is a justified conclusion (it is necessary to go through the whole process of constructing arguments and establishing extensions to determine this). 

From this we conclude that though $\vsimj$ is not cumulative, and hence \aspicplus\ is, in some sense, weaker than non-monotonic logics like circumscription \cite{mccarthy1980circumscription} and default logic \cite{reiter1980logic}, it is not clear that it is too weak. That is strengthening $\vsimj$ so that it is cumulative for defeasible theories would allow conclusions that make no sense from the point of view of argumentation-based reasoning. Whether there are other ways to strengthen \aspicplus\ that do make sense is an open question, and one we will investigate.

\section{The Rationality Postulates}
\label{sec:rationality}

Finally, we consider the three postulates of \cite{caminada2007evaluation} (which \aspicplus\ complies with), namely (1) closure under strict rules; and (2) direct and (3) indirect consistency.
We ask whether the axioms discussed in this paper are equivalent to any of these postulates. In what follows, we assume that strict rules are consistent.

\subsection{Closure under strict rules}
\begin{proposition}
An argumentation framework meets closure under strict rules if and only if  the consequence relation for strict rules complies with right weakening (RW) with regards to justified conclusions.
\begin{proof}
Given an argumentation framework $AF$, assume that $\alpha$ is in the justified conclusions. Therefore $\top \vsim_j \alpha$, and assume that there is a strict rule $\models \alpha \to \beta$. Using RW, we obtain $\top \vsim_j \beta$. Therefore RW implies closure under strict rules. Furthermore, as per the proof of Proposition~\ref{prop:justified}, closure under strict rules implies that $\vsim_j$ satisfies RW.
\end{proof}
\end{proposition}

\subsection{Direct consistency}
Direct consistency with regards to $\vsim_j$ requires that no extension contains inconsistent arguments (and therefore inconsistent conclusions). This is equivalent to the following axiom,  unobtainable from the axioms discussed previously. 
\[
\begin{bprooftree}
\AxiomC{$\alpha \vsim_j \beta$}
\UnaryInfC{$\alpha \not\vsim_j \overline{\beta}$}
\end{bprooftree}
\]

\subsection{Indirect Consistency}
\begin{proposition}
Assume we have direct consistency, and that strict rules are consistent. Any system which satisfies monotonicity under strict rules will satisfy indirect consistency, and vice-versa.
\begin{proof} From  \cite[Prop.~7]{caminada2007evaluation},  direct consistency and closure yield indirect consistency. We assume direct consistency, and monotonicity gives closure.
\end{proof}
\end{proposition}

In this section we have shown that the rationality postulates described in \cite{caminada2007evaluation} can be described using axioms from classical logic and non-monotonic reasoning. In future work, we intend to determine whether these axioms can help identify additional rationality postulates. In addition, we will investigate whether these axioms can represent the additional rationality postulates described in \cite{wu12between}.

\section{Related Work}
\label{sec;related}
There are several papers describing work that is similar in some respects to what we report here.
Billington \shortcite{billington1993defeasible} describes Defeasible Logic, a logic that, as its name implies, differs from classical logic in that it deals with defeasible reasoning. In addition to introducing the logic, \cite{billington1993defeasible} shows that defeasible logic satisfies the axioms of reflexivity, cut and cautious monotonicity suggested in \cite{gabbay:nato}, thus satisfying what \cite{gabbay:nato} describes as the basic requirements for a non-monotonic system (such a system is equivalent to a cumulative system in \cite{kraus1990nonmonotonic}). \cite{governatori2004argumentation} subsequently established significant links between reasoning in defeasible logic and argumentation-based reasoning. To do this, \cite{governatori2004argumentation} provides an argumentation system that makes use of defeasible logic as its underlying logic, and shows that the system is compatible with Dung's semantics \cite{phan1995acceptability}. Given Defeasible Logic's close relation to Prolog \cite{nute2001defeasible}, this line of work is closely related to Defeasible Logic Programming (DeLP) \cite{garcia2004defeasible}, a formalism combining results of Logic Programming and Defeasible Argumentation. As a rule-based argumentation system, DeLP also has strict/defeasible rules and a set of facts. DeLP differs from \aspicplus\ in the types of attack relation it permits (no undermining) and in the way that it computes conclusions (it does not implement Dung's semantics). 

\section{Conclusions}
\label{sec:conclusion}
In this paper we considered which of the axioms of \cite{kraus1990nonmonotonic} \aspicplus\ meets based on two different interpretations of the consequence relation. 
Somewhat surprisingly, we found that when strict and defeasible elements are both present, \aspicplus\ is not cumulative, and even in the presence of only strict rules, is not monotonic. 
These results suggest that additional constraints need to be placed on any instantiation of \aspicplus. 
Since \aspicplus\ is intended as a template that can be instantiated with other logics (by the inclusion of strict rules that capture the rules of inference of those logics), this is perhaps unsurprising.
However, it does place constraints on what those logics must bring if certain properties are required of that instantiation of \aspicplus.
For example, rules that capture contraposition are required for the strict part of \aspicplus\ to be monotonic in the sense of \cite{kraus1990nonmonotonic}.
We also investigated the relationship between the axioms of \cite{kraus1990nonmonotonic} and the rationality postulates, and suggested an alternative, axiom based formulation of the latter.

As mentioned above, we will investigate whether additional axioms can encode the rationality postulates described in \cite{wu12between}. We will also examine the properties of different interpretations of the logical symbols. For example, we assumed that $\equiv$ encodes the presence of two rules, but says nothing about their preferences or defeaters. Finally, we may consider other interpretations of the consequence relation. This paper therefore opens up several significant avenues of future investigation.

\newpage
\bibliographystyle{plain}
\bibliography{nonmonotonic}

\end{document}